\pdfoutput=1
\documentclass{article}
\usepackage{spconf,amsmath,graphicx}
\usepackage{amssymb}
\usepackage{booktabs}
\usepackage{multirow}
\usepackage[caption=false]{subfig}
\usepackage{siunitx}
\usepackage[export]{adjustbox}
\usepackage{array}
\newcommand{\PreserveBackslash}[1]{\let\temp=\\#1\let\\=\temp}
\newcolumntype{C}[1]{>{\PreserveBackslash\centering}p{#1}}
\newcolumntype{R}[1]{>{\PreserveBackslash\raggedleft}p{#1}}
\newcolumntype{L}[1]{>{\PreserveBackslash\raggedright}p{#1}}
\usepackage{enumitem}
\usepackage{xcolor}

\title{The arm-swing is discriminative in video gait recognition \\ for athlete re-identification }
\name{Yapkan Choi, Yeshwanth Napolean, Jan C. van Gemert\thanks{This study is supported by NWO grant P16-28 project 8 Monitor and prevent thermal injuries in endurance and Paralympic sports of the Perspectief Citius Altius Sanius - Injury-free exercise for everyone program.}}
\address{Computer Vision Lab, Delft University of Technology, Delft, The Netherlands}
\begin{document}
%
\maketitle
\begin{abstract}
In this paper we evaluate running gait as an attribute for video person re-identification in a long-distance running event. 
We show that running gait recognition achieves competitive performance compared to appearance-based approaches in the cross-camera retrieval task and that gait and appearance features are complementary to each other.
For gait, the arm swing during running is less distinguishable when using binary gait silhouettes, due to ambiguity in the torso region.
We propose to use human semantic parsing to create partial gait silhouettes where the torso is left out.
Leaving out the torso improves recognition results by allowing the arm swing to be more visible in the frontal and oblique viewing angles, which offers hints that arm swings are somewhat personal.
Experiments show an increase of $3.2\%$ mAP on the CampusRun and increased accuracy with $4.8\%$ in the frontal and rear view on CASIA-B, compared to using the full body silhouettes. 
\end{abstract}
\section{Introduction}
\label{sec:introduction}
Athletes in long-distance running events are typically identified and tracked using the number tag on their race bib, which may include a RFID tag for measuring split times at specific locations, or a GPS tracker for real-time tracking.
In smaller events, usually only the start and finish time are registered, not intermediate locations and times.
With increasing prevalence of smartphones/cameras, images and videos from race organizers or spectators provide an additional source of information for runner identification and tracking \cite{Flintham2015}.
Vision-based methods for identifying distance runners include bib number detection \cite{Ben-Ami, Shivakumara2017, Boonsim2018, karaoglu2017text} and appearance-based person re-identification \cite{Napolean2019a}.
Potential issues with these methods arise when the bib number is partially or fully occluded, or when multiple athletes wear similar clothing styles and color. Additionally, avoiding the storage of RGB images would alleviate privacy concerns where people can easily be recognized by others. In this paper we investigate if identifying runners based on their running gait is possible and we explore gait recognition as a complementary alternative to runner re-identification with appearance features.

\begin{figure}[t]
\begin{minipage}[b]{1.0\linewidth}
  \centering
  \centerline{\includegraphics[width=8.5cm]{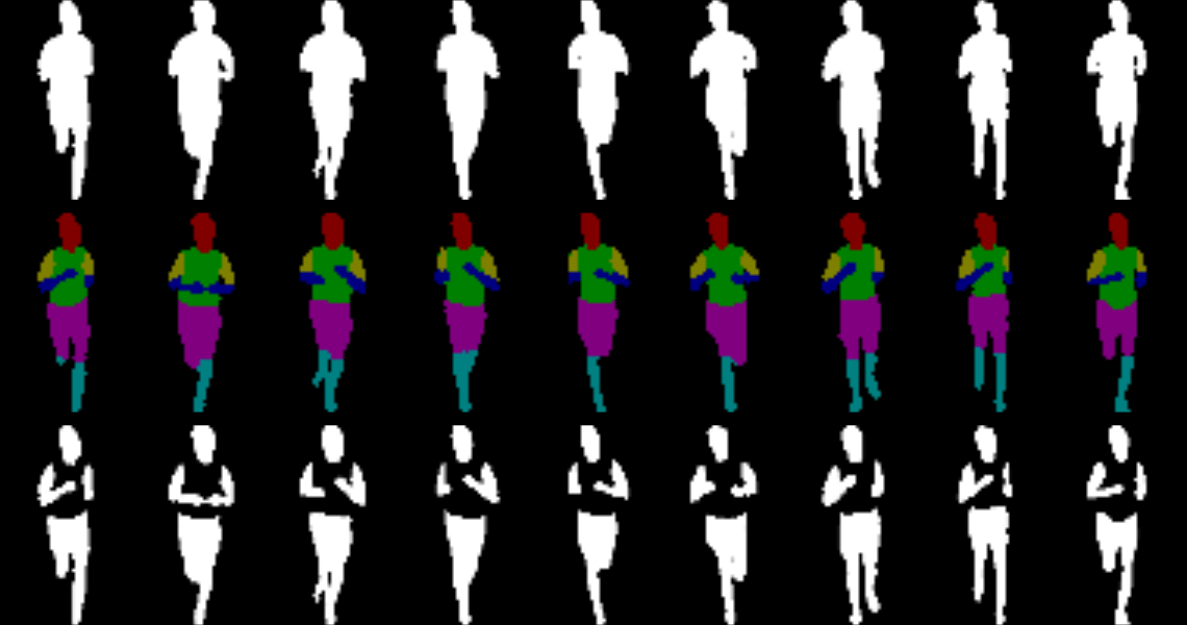}}
\end{minipage}
\caption{With binary gait silhouettes, the arm swing is not visible (top). We reduce the ambiguity caused by self-occlusion with the torso with body-part-specific segmentation masks (middle). Partial binary gait silhouettes of a running gait cycle with visible arm swing by removing the torso (bottom).}
\label{fig:body-part-specific}
\end{figure}

Recently, research in gait recognition has focused on dealing with co-variates such as view angle \cite{Wu2017, Takemura2018}, clothing and carrying conditions \cite{Yu2006}. 
Although speed-invariant gait recognition from treadmill sequences has been proposed before \cite{Guan2013, Xu2019a}, to the best of our knowledge, no previous research has evaluated running gait recognition with unconstrained running conditions.
In this paper, we use the CampusRun dataset~\cite{Napolean2019a} of videos captured by hand-held cameras during a running event and evaluate models in a cross-camera setting.

Representing gait as a sequence of binary gait silhouettes has been widely adopted \cite{Wu2017, Uddin2019, Chao2019, Fan2020}. The primary concern with silhouettes for running gait is self-occlusion of body parts. Specifically, a portion of the arm swing is lost due to the ambiguity of the torso region when using gait silhouettes. As the arm swing is informative, we propose to create partial silhouettes (from binary images) from body-part-specific segmentation masks generated by a  semantic body-part parsing model \cite{Li2019a}.
The partial binary gait silhouettes without the torso reduces the ambiguity of the torso region and improves person re-identification results by allowing the arm swing to be much more visible (see figure \ref{fig:body-part-specific}). 
Additionally, we evaluate the partial silhouettes on walking sequences.
As we walk with straight arms, ambiguity in the binary gait silhouettes is less prevalent.
Our main contributions can be summarized as follows:
\begin{itemize}[leftmargin=*,align=left]
\item We apply cross-camera video person re-identification in the long-distance running domain by extending the CampusRun video dataset \cite{Napolean2019a} with 2,581 annotated tracklets of 257 recreational runners from 18 cameras.

\item We compare and complement gait features with appearance features on the CampusRun.
We demonstrate the feasibility and usefulness of gait as a feature for the cross-camera retrieval task.

\item We show that removing the torso from the silhouettes provides $3.2\%$ improved mAP on the CampusRun and we also achieved a $4.8\%$ increase in performance with the CASIA-B dataset (frontal and rear view).
\end{itemize}
\section{Method}
\label{sec:method}
\subsection{Gait silhouette}
\label{subsec:method-gait}
Given a sequence of bounding boxes indicating a subjects position in an image/video frame (tracklet) of a runner, we construct silhouettes using bounding boxes from consecutive frames.
Background subtraction \cite{Wang2003} for extracting gait silhouettes requires a static camera for reliable results.
Since we allow hand-held cameras, we use convolutional neural networks for segmenting gait silhouettes from the tracklets.

\begin{figure}[t]
\begin{minipage}[b]{1.0\linewidth}
  \centering
  \centerline{\includegraphics[width=8.5cm]{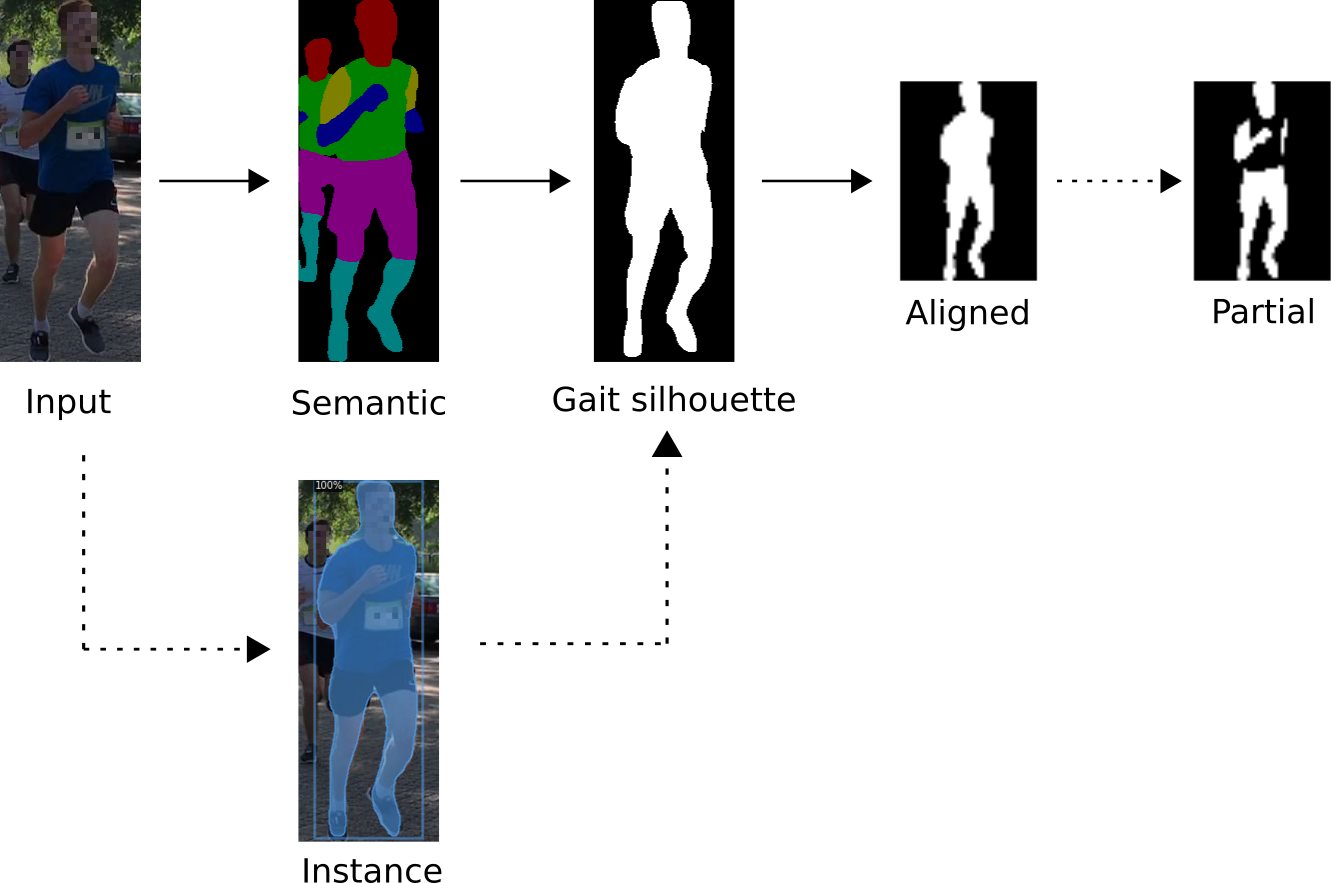}}
\end{minipage}
\caption{Our pipeline with human semantic parsing \cite{Li2019a} and instance segmentation \cite{He2017a} to create (partial) binary gait silhouettes from bounding boxes. Instance segmentation is only used when there are multiple persons in the bounding box.}
\label{fig:sil-pipeline}
\end{figure}

\textbf{Pipeline.}
Figure \ref{fig:sil-pipeline} depicts the pipeline for constructing binary gait silhouettes.
For each bounding box, we use a human semantic parsing model \cite{Li2019a} to segment the input images into body-part-specific masks. 
As the human parsing model is on a semantic level and the bounding box can contain multiple identities, we use Mask R-CNN \cite{He2017a} to segment the person of interest and keep only the largest instance when multiple instances are found by Mask R-CNN.
The body-part-specific masks are converted to binary silhouettes, aligned, and resized to a size of 64$\times$44, following the GaitSet approach \cite{Chao2019}.

\textbf{Partial silhouettes.}
The human semantic parsing model \cite{Li2019a} in the pipeline is pre-trained on the PASCAL-Person-Part dataset \cite{Chen2014}.
Unlike other human semantic parsing datasets \cite{Liang2015, Gong2017}, the PASCAL-Person-Part dataset does not have clothing-specific segmentation label categories.
We use 7 labels: Background, Head, Torso, Upper Arms, Lower Arms, Upper Legs and Lower Legs.
These body-part-specific labels suit the task of gait recognition, because the resulting segmentation masks are less dependent on the person's clothing.
The partial gait silhouettes are composed of all body-part-specific segmentation masks without the torso.

\subsection{Models.}
We use a baseline gait recognition model and two appearance-based person re-identification models to compare gait and appearance features.
For a fair comparison, all models use identical input sampling, input resolution and loss function.

\begin{figure*}[ht]
\small
\centering
	\subfloat[Sample]{\includegraphics[height=0.1\linewidth,valign=t]{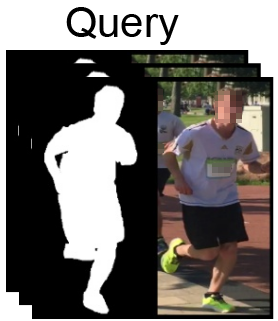}}
	\hfil
	\subfloat[GaitSet feature]{\includegraphics[height=0.1\linewidth,valign=t]{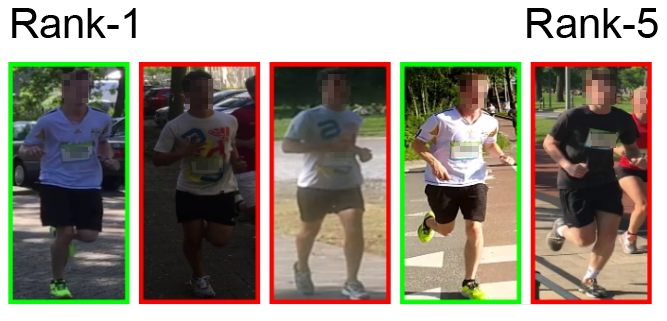}}
	\hfil
	\subfloat[2D ResNet-50 feature]{\includegraphics[height=0.1\linewidth,valign=t]{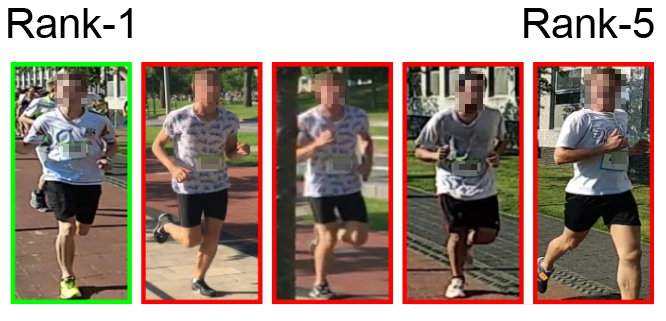}}
	\hfil
	\subfloat[Combined feature]{\includegraphics[height=0.1\linewidth,valign=t]{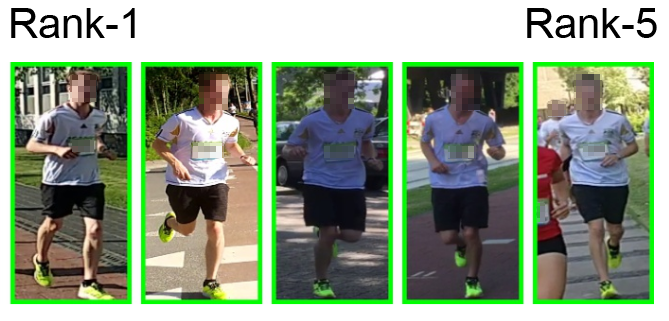}}
\caption{(a) Sample query and corresponding rank-5 retrieval results for (b) GaitSet, (c) 2D ResNet-50 and (d) their combined feature. 
Green and red borders denote correct and incorrect matches respectively.
Gait features (b) are more robust against co-variates such as background and clothing color. Multi-modal approach (d) improves retrieval results.}
\label{fig:sample-result}
\end{figure*}

\textbf{Gait features.}
We use GaitSet \cite{Chao2019} as our baseline gait recognition model. 
It achieves state-of-the-art performance on CASIA-B \cite{Yu2006} and OU-MVLP \cite{Takemura2018} for cross-view gait recognition.
In GaitSet, the identity of a person is learned from a set of gait silhouettes. 
The network first extracts frame-level features and then aggregates the feature maps of each silhouette using max pooling at the set-level.
Horizontal Pyramid Pooling \cite{Fu2018} slices the last set-level feature map into different horizontal strips of multiple pyramid scales, to learn feature representations with different receptive fields and spatial locations.
For each set of silhouettes, the network outputs a discriminative representation, consisting of 62 feature map strips with 256 dimensions each.
During training, the set of silhouettes is a subset of the sequence, where we randomly sample a fixed number of silhouettes from the tracklet.
As human gait is a periodic movement, a representation can be learned if we sample sufficient frames.
All silhouettes from the tracklet are used during evaluation.

\textbf{Appearance features.}
For appearance-based person re-identification models, we explore 2D and 3D CNN models with a ResNet-50 backbone \cite{He2016}.
Like our baseline gait recognition model, both appearance-based models use a randomly sampled subset of bounding boxes during training.
For evaluation, both models output a feature vector with 2,048 dimensions for each input tracklet.
We use a 2D ResNet-50 \cite{He2016} model, pre-trained on Image{N}et \cite{Russakovsky2015}. 
The model aggregates frame-level features using average pooling to get one feature representation for the set of input bounding boxes.
To leverage features from both the temporal and spatial dimensions, we use a 3D ResNet-50 \cite{Hara2018} model which is pre-trained on Kinetics \cite{Kay} for the action recognition task.
In contrast to GaitSet and 2D ResNet-50, we use randomly sampled sequences with consecutive frames for 3D ResNet-50 during training.
We use the layer before the final classification layer as the person identity feature.
During testing, a tracklet gets split into non-overlapping chunks with a fixed number of consecutive frames, followed by taking the mean of the person identity features from each chunk.

\textbf{Triplet loss.}
Models are trained with Batch-All triplet loss \cite{Hermans2017}, where all triplet combinations in a batch are used for calculating the loss.
The triplet loss in GaitSet is calculated for each of the 62 feature strips individually, followed by taking the mean of the losses.
The batch size is $p \times k \times c$, where $p$ denotes the number of people, $k$ the number of tracklets for each person and $c$ the number of frames for each tracklet.
\section{Experiments}
\label{sec:experiments}

\subsection{Comparison with appearance-based features}
\label{subsec:experiment1}
\textbf{CampusRun dataset.}
The CampusRun \cite{Napolean2019a} was a running event with 257 runners who were captured on video using 9 non-stationary hand-held smartphone cameras across the whole track, where each camera operator was allowed to move along the course.
We use multi-object tracking \cite{Zhang2020} to extract tracklets and bounding boxes for each runner from the videos.
After manually annotating the bib numbers, we obtain bounding boxes for 257 runners and 2,581 tracklets with an average sequence length of 77 frames.
The 10 km runners have 13 tracklets on average.

\textbf{Evaluation protocol.}
We use the 5 km runners for model training and validation, while the 10 km runners are only used for testing.
The training set and validation set are constructed using a 60/40 split (5 km, 125 runners, 9 cameras, 860 tracklets).
The test set (10 km, 132 runners, 18 cameras, 1,721 tracklets) and validation set are evaluated using a cross-camera setting, where the probe identity is captured from a different camera than the positive matches in the gallery.
We have 1,721 test queries with a maximum of 17 positive matches for each query, as the runners do not appear more than once per camera.

\textbf{Training details.} 
We follow the training protocol of GaitSet \cite{Chao2019} for all models, but use a smaller batch size ($p=8$ persons, $k=4$ tracklets) due to the CampusRun dataset having fewer sequences per identity than in CASIA-B.
Additionally, the GaitSet model is pre-trained on CASIA-B.
The learning rate is set to 1e-4, and we train the models for 80K iterations.
We choose the best model checkpoint based on the mAP of the validation set.
During training, we randomly sample $c=30$, $c=10$ and $c=30$ frames for GaitSet, 2D ResNet-50 and 3D ResNet-50 respectively.
For data augmentation, we randomly horizontal flip the entire tracklet.
We compare the output vectors of two tracklets using Euclidean distance.
We resize and align the silhouettes to 64$\times$44, while the bounding boxes for the 2D ResNet-50 and 3D ResNet-50 are resized to 64$\times$32.

\begin{table}[tbp]
\small
\centering
\begin{tabular}{@{}lccc@{}} \toprule
Method & mAP & Rank-1 \\
\midrule
Appearance: 2D ResNet-50 \cite{He2016} & \textbf{56.3} & 74.6 \\
Appearance: 3D ResNet-50 \cite{Hara2018} & 56.2 & 74.0 \\
Gait: GaitSet \cite{Chao2019} & 52.2 & \textbf{78.7} \\
\midrule
GaitSet \cite{Chao2019} + 2D ResNet-50 \cite{He2016} & \textbf{81.1} & \textbf{93.9} \\ 
GaitSet \cite{Chao2019} + 3D ResNet-50 \cite{Hara2018} & 71.1 & 90.1 \\ 
2D ResNet-50 \cite{He2016} + 3D ResNet-50 \cite{Hara2018} & 59.3  & 78.4 \\
\bottomrule
\end{tabular}
\caption{\textbf{Exp. 1:}
Comparison of appearance-based and gait-based methods on CampusRun.
Gait features achieves comparable mean average precision and rank-1 accuracy to appearance-based methods.
The pairwise model combinations show that gait features are complementary to appearance features for person re-identification.}
\label{tab:video-based}
\end{table}

\textbf{Exp. 1: Results on CampusRun.} 
Table \ref{tab:video-based} shows the comparison between appearance-based and gait-based models on the CampusRun dataset.
We observe comparable mAP and rank-1 accuracy between all three models.
We use pairwise combinations of the three models to analyze if the models learn different features.
Before performing distance calculations, we concatenate the two feature vectors from each pair of models, after first $\ell_2$ normalizing the individual vectors.
The results for the pairwise combinations in table \ref{tab:video-based} show that a multi-modal approach using gait and appearance features leads to a more diverse and complementary ensemble than adding a spatio-temporal model from the same modality.
Retrieval results in figure \ref{fig:sample-result} highlight advantages and disadvantages of the respective methods.

\subsection{Evaluating partial binary gait silhouettes}
\label{subsec:experiment2}
\textbf{Datasets.}
For this experiment, we train the GaitSet model from scratch using partial gait silhouettes without the torso. 
We use the CampusRun dataset as described in section \ref{subsec:experiment1}.
Additionally, we explore partial gait silhouettes for walking sequences using the CASIA-B \cite{Yu2006} dataset.
It contains gait sequences of 124 persons with 3 walking conditions: normal (6 sequences NM\#1-6), carrying a bag (2 sequences BG\#1-2) and wearing a coat (2 sequences CL\#1-2). 
The participants are captured from 11 views from 0\si{\degree} to 180\si{\degree} in 18\si{\degree} increments, resulting in $11 \times (6 + 2 + 2) = 110$ sequences for each person.

\textbf{Evaluation protocol.}
We follow the same setup and evaluation protocol as in GaitSet \cite{Chao2019}.
The first 74 persons are used for training and the remaining 50 persons for testing.
The models are evaluated with rank-1 accuracy, but we exclude identical-view cases.

\textbf{Exp. 2: Results on CampusRun.}
Table \ref{tab:pipeline} shows the contribution of each component of our proposed pipeline for constructing partial gait silhouettes.
The segmentation masks generated by the human semantic parsing model \cite{Li2019a} are more detailed than segmentation masks from the instance segmentation model \cite{He2017a}, leading to a $13.9\%$ improvement in mAP.
Partial gait silhouettes without the torso increases the portion of arm swing that is visible, resulting in a $3.2\%$ mAP improvement over using the full body silhouettes.

\begin{table}
\setlength\tabcolsep{3.0pt}
\small
\centering
	\begin{tabular}{@{}ccccc@{}} \toprule
		Instance seg. & Semantic parsing & Partial silhouettes & mAP & Rank-1 \\
		\midrule
		\checkmark & & & 37.4 & 64.9 \\
		\checkmark & \checkmark & & 51.3 & 78.2\\
		\checkmark & \checkmark & \checkmark & \textbf{54.5} & \textbf{81.1} \\
		\bottomrule
	\end{tabular}
\caption{\textbf{Exp. 2:} The contribution of each component of our proposed pipeline towards performance on CampusRun. The largest improvement is attributed to the increase in detail of human semantic parsing silhouettes.}
\label{tab:pipeline}
\end{table}

\textbf{Exp. 3: Results on CASIA-B.}
The torso segmentation mask is subtracted from the original binary gait silhouettes provided by CASIA-B. 
Table \ref{tab:casia-b-parts} shows the average rank-1 accuracies for full body silhouettes and partial silhouettes without the torso.
For all models and co-variates, oblique view angles (18\si{\degree}-72\si{\degree}, 108\si{\degree}-162\si{\degree}) achieve higher accuracy than frontal (0\si{\degree}, 180\si{\degree}) or lateral views (90\si{\degree}), because silhouettes observed from oblique view angles contain more motion information than the other two planes individually.
Subtracting the torso from the original silhouettes  increases accuracy for the frontal views, because contours of the arm swing become more perceptible.
With binary gait silhouettes, it is difficult to discern the human gait in the frontal plane, as the arm and leg swing happen in the sagittal plane.
Without the torso, the motion in the frontal plane is more visible, which improves accuracy for the frontal results.
A similar pattern of results was obtained for the bag and clothing co-variates.

For CASIA-B, it remains unclear to which degree recognition performance is attributed to the pixel-level accuracy of the segmentation masks generated by the human semantic parsing model.
We observe incorrect parsing results for the lateral view angle when the arms align with the torso.
This may explain why the accuracy for the lateral view angle decreases for all three probe subsets (NM, BG, CL), when subtracting the torso from the silhouettes.
Most sequences in the CampusRun dataset were captured from oblique angles between 10\si{\degree} and 45\si{\degree}.
We did not find an increase in rank-1 accuracy for oblique angles in CASIA-B as was observed in the CampusRun, which suggests that the arm swing is more discriminative for recognizing running gait.

\begin{table}[t]
\setlength\tabcolsep{2.4pt}
	\small
	\centering
	\begin{tabular}{@{}lllcccc@{}} \toprule
		\multicolumn{3}{l}{Gallery NM\#1-4} & \multicolumn{3}{c}{0\si{\degree}-180\si{\degree}} & \\
		\midrule
		Probe & Method & Silhouette & Frontal & Oblique & Lateral & Mean\\
		\midrule
		\multirow{3}{*}{NM\#5-6} & GaitPart \cite{Fan2020} & Full body & 92.3 & \textbf{97.7} & \textbf{92.3} & \textbf{96.2} \\
		& GaitSet \cite{Chao2019} & Full body & 88.3 & 97.1 & 91.7 & 95.0 \\
		& GaitSet \cite{Chao2019} & Partial & \textbf{93.1} & 97.1 & 91.1 & 95.8 \\
		\midrule
		\multirow{3}{*}{BG\#1-2} & GaitPart \cite{Fan2020} & Full body & \textbf{87.5} & \textbf{93.4} & \textbf{84.9} & \textbf{91.5} \\
		& GaitSet \cite{Chao2019} & Full body & 81.4 & 89.5 & 81.0 & 87.2 \\
		& GaitSet \cite{Chao2019} & Partial & 82.3 & 89.6 & 79.1 & 87.3\\
		\midrule
		\multirow{3}{*}{CL\#1-2}& GaitPart \cite{Fan2020} & Full body & \textbf{68.6} & \textbf{82.0} & \textbf{72.5} & \textbf{78.7} \\
		& GaitSet \cite{Chao2019} & Full body & 55.7 & 74.1 & 70.1 & 70.4 \\
		& GaitSet \cite{Chao2019} & Partial & 62.2 & 75.4 & 67.0 & 72.2 \\
		\bottomrule
	\end{tabular}
	\caption{\textbf{Exp. 3:}
	Averaged rank-1 accuracies on CASIA-B.
		The oblique (18\si{\degree}-72\si{\degree}, 108\si{\degree}-162\si{\degree}) and frontal (0\si{\degree}, 180\si{\degree}) probe views are grouped together in the table, but the mean accuracy is calculated over all 11 views.
		Subtracting the torso leads to increased accuracy in the frontal views, but decreased accuracy in the lateral view.
	}
	\label{tab:casia-b-parts}
\end{table}
\section{Conclusion}
\label{sec:conclusion}
We extend the CampusRun dataset \cite{Napolean2019a} with additional annotations, recorded at a long-distance running event for cross-camera video person re-identification. 
Experimental results using the CampusRun dataset show that runners can be identified based on their running gait.
Furthermore, we demonstrate that gait features are both competitive and complementary to appearance features.
Additionally, we investigate arm swing as a feature by extracting partial binary gait silhouettes using human semantic parsing and instance segmentation. 
We demonstrate that subtracting the torso from the gait silhouettes for runners leads to increased recognition performance by making the arm swing more visible. 
\bibliographystyle{IEEEbib_short}
\bibliography{library}

\end{document}